\documentclass{article}
\usepackage{spconf, amsmath, graphicx, booktabs, float, hyperref}

\title{Decoding Emotions: A comprehensive Multilingual Study \\of Speech Models for Speech Emotion Recognition}

\twoauthors
 {Anant Singh}
	{New York University\\
	New York, USA}
 {Akshat Gupta}
	{JP Morgan Chase\\
	New York, USA}

\begin{document}
\maketitle

\begin{abstract}
Recent advancements in transformer-based speech representation models have greatly transformed speech processing. However, there has been limited research conducted on evaluating these models for speech emotion recognition (SER) across multiple languages and examining their internal representations. This article addresses these gaps by presenting a comprehensive benchmark for SER with eight speech representation models and six different languages. We conducted probing experiments to gain insights into inner workings of these models for SER. We find that using features from a single optimal layer of a speech model reduces the error rate by 32\% on average across seven datasets when compared to systems where features from all layers of speech models are used. We also achieve state-of-the-art results for German and Persian languages.  Our probing results indicate that the middle layers of speech models capture the most important emotional information for speech emotion recognition. GitHub\footnote{https://github.com/95anantsingh/Decoding-Emotions} 

\end{abstract}

\begin{keywords}
Speech Emotion Recognition, HuBERT, wav2vec2, Edge Probing, Feature Extraction.
\end{keywords}

\section{Introduction}
\label{sec:introduction}

In recent years, several transformer-based speech representation models, trained on massive amounts of speech data, have been introduced \cite{baevski2020wav2vec} \cite{conneau2020unsupervised} \cite{hsu2021hubert} \cite{radford2023robust}. These models have demonstrated remarkable improvements in performance across various downstream tasks. Although several benchmarks exist for evaluating their performance \cite{yang2021superb} \cite{shi2023ml}, insufficient attention has been given to evaluating these models across multiple languages. Previous large-scale works on speech emotion recognition either do not evaluate the performance of the latest speech representation models \cite{scheidwasser2022serab} or they only evaluate a subset of models or languages \cite{pepino2021emotion} \cite{sharma2022multi} \cite{wang2023multilingual}. Consequently, there is a gap in the literature when it comes to comprehensive evaluations of these models for speech emotion recognition across various languages.

Another critical aspect that has received limited attention is analyzing the inner workings of these models using probing techniques, which can provide valuable insights into how these models process and encode various linguistic and acoustic features. By examining the responses of these models to specific linguistic or emotional cues, we can develop a deeper understanding of the strengths and limitations of these models. Recognizing emotion in speech requires an understanding of both phonetic and the prosodic content in the spoken utterance \cite{lin2023utility}. While previous work has explored probing techniques to understand the phonetic content of speech representation models \cite{shah2021all} \cite{pasad2021layer} \cite{pasad2023comparative}, we haven't seen studies doing the same for tasks that have a higher dependence on prosodic content.

This paper has a dual focus. Firstly, we present a comprehensive benchmark of multiple speech representation models for speech emotion recognition across a range of languages. This benchmark ensures that the models are just as applicable and relevant in diverse cultural and linguistic contexts. One of the problems in SER is the lack of standardized training and testing splits which allows different papers to report different performances for the task. To counter this, we adopt a standardized train-dev-test split of Scheidwasser et al. (2022) \cite{scheidwasser2022serab} to facilitate consistent comparisons. By providing a standardized evaluation framework, we enable effective comparisons and assessments of the performance of different speech emotion recognition models.

\begin{figure*}[t]
  \centering
    \includegraphics[width=0.99\textwidth]{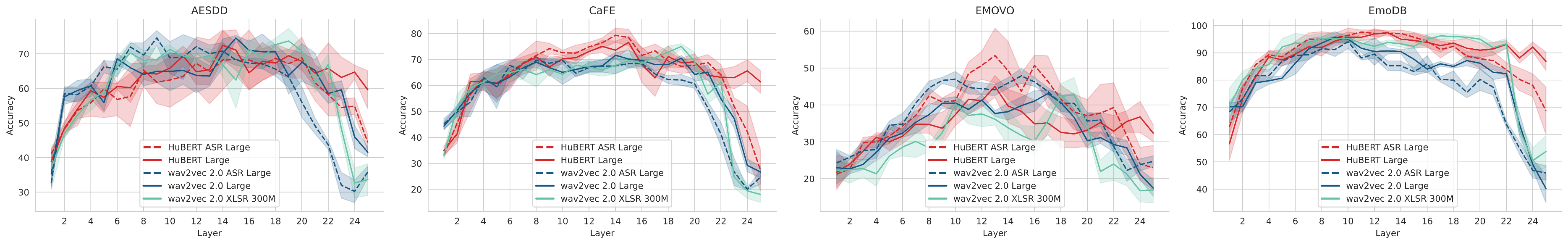}
    \includegraphics[width=0.99\textwidth]{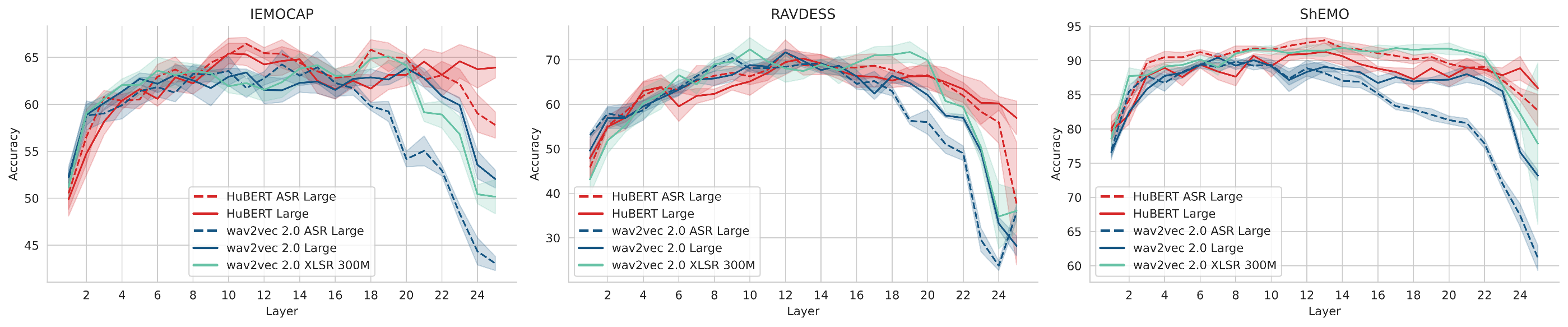}
  \caption{Dataset-wise accuracy for Probing using a \textit{Linear} classifier with various feature extractors. The presented data is an average of five runs conducted on the test set.}
  \label{fig:linear_acc_plots}
\end{figure*}

Secondly, we conduct probing experiments to gain insights into the underlying mechanisms of speech emotion recognition across multiple languages. Unlike \cite{lin2023utility}, we find that the most important layer for speech emotion recognition are the center layers, as can be seen in Figure \ref{fig:linear_acc_plots}. This observation then inspired us to just utilize one single layer instead of the final layer or aggregate all layers \cite{pepino2021emotion} for SER. We surprisingly find that when features are extracted from the optimal layer, which are usually the center layers of the model, we achieve best performance. This is contrary to previous work where conventionally only final layer features or aggregating feature from different layers \cite{pepino2021emotion} have lead to best results. As a result, we report state-of-the-art results for German and Persian where only speech input is used for SER. Our findings challenge the prevailing notion and underscore the importance of selecting the appropriate layer to maximize the efficacy of speech representation models in SER tasks.

\begin{table*}[t]
  \centering
  \begin{tabular}{llccccccc}
    \toprule
    Dataset  & Language & Classes & Utterances & Speakers & Average Duration (s) & Total Duration (h) \\
    \midrule
    AESDD   & Greek    & 5 & 604   & 6     & 4.2 & 0.7 \\
    CaFE    & French   & 7 & 864   & 12    & 4.5 & 1.1 \\
    EmoDB   & German   & 7 & 535   & 10    & 2.8 & 0.4 \\
    EMOVO   & Italian  & 7 & 588   & 6     & 3.1 & 0.5 \\
    IEMOCAP    & English  & 4 & 5,531 & 10    & 3.4 & 7.0 \\
    RAVDESS & English  & 8 & 1,440 & 24    & 3.7 & 1.5 \\
    ShEMO   & Persian  & 6 & 3,000 & 87    & 4.0 & 3.3 \\
    \bottomrule
  \end{tabular}
  \caption{Key Attributes of Datasets Used: Average Duration, Language, Number of Classes, Utterances, Speakers, and Total Duration of All Samples}
    \label{tab:datasets}
\end{table*}

\section{Background}
Previous research on probing speech representation models have primarily focused on studying how speech models understand phonetic content \cite{shah2021all} \cite{pasad2021layer} \cite{pasad2023comparative}. In order to probe each layer, these studies adopt a technique of obtaining the time-average of feature vectors from the target layer and then passing them through a linear layer for downstream tasks. The idea is to use the least complex model for classification so that the descriptive power of the model can be studied. Notably, these investigations reveal that wav2vec2 exhibits an auto-encoder-like behavior, with the initial and final layers resembling the input, while the intermediate layers generate higher-level representations that encapsulate maximum contextual information. In a study conducted by Lin et al. (2023) regarding the utilization of speech representation models for prosodic tasks, it was discovered that only the initial layers encoded prosodic information. We find that speech emotion recognition is a task done best by the center layers of the model, producing state-of-the-art results with a very simple linear model when applied on the right layer. 

A large number of research articles exists in literature for making systems for speech emotion recognition. In this paper, we focus on doing speech emotion recognition with transformer-based speech representation models. We specifically focus on using three pre-trained speech models - wav2vec2 \cite{baevski2020wav2vec}, XLSR \cite{conneau2020unsupervised} and HUBERT \cite{hsu2021hubert}. We do not fine-tune the weights of the speech representation models \cite{wang2021fine} \cite{sharma2022multi}\cite{chen2022does}\cite{chen2023exploring} and only use them as feature extractors as we believe such systems are more practical in the real world setting - where a single feature extractor is used to extract speech features and multiple systems perform different tasks using the same set of features. Unlike pre-trained language models in natural language processing where the final layer features are used to do most tasks, Pepino et al \cite{pepino2021emotion} showed that the best way to use speech models for downstream task is not to use the final layer representations. They took a weighted averaged the features of across all layers of wav2vec2 and then used an LSTM model to achieve state-of-the-art results for the IEMOCAP (English) dataset. In our paper, we show that better results can be achieved using information from the optimal layer. 

In this paper, we also study speech emotion recognition across multiple languages. Most studies on speech emotion are centred around English \cite{pepino2021emotion} \cite{cai2021speech} \cite{li2023exploration} or use one or two additional languages 
\cite{yazdani2021emotion} \cite{wang2023multilingual} \cite{sadok2023vector}. Our work also presents a benchmark for speech emotion recognition across multiple languages using pre-trained speech representation models. Previous multlingual SER works either don't evaluate these representation models \cite{scheidwasser2022serab} or only use one such model for multiple languages \cite{sharma2022multi}. While such benchmarks exist in low-resourced languages \cite{gupta2022building} \cite{gupta2021acoustics} \cite{gupta2021intent} for other tasks \cite{gupta2021intent} \cite{shi2023ml}, we do not yet have a comprehensive benchmark for speech emotion recognition.

\section{Datasets}
\label{sec:datasets}

This research explores seven speech emotion classification tasks conducted on six distinct language datasets. The study focuses on English, French, German, Greek, Italian, and Persian. Specifically, the English datasets used in this study consist of two widely recognized collections, namely IEMOCAP (IEM4) \cite{IEMOCAP} and RAVDESS \cite{RAVDESS}. Additionally, the other languages are represented by CaFE (French) \cite{CaFE}, EmoDB (German) \cite{EmoDB}, AESDD (Greek) \cite{AESDD}, EMOVO (Italian) \cite{EMOVO}, and ShEMO (Persian) \cite{ShEMO} dataset.

The datasets exhibit variations in several aspects, including size (number of utterances), number of speakers, class distribution, and number of classes as described in Table \ref{tab:datasets}. While emotions such as anger, happiness, and sadness are present across all datasets, additional emotions such as disgust, fear, neutral emotion, surprise, calm, and boredom appear in at least one of the datasets. Each dataset consists of speech samples characterized by three crucial attributes: audio data represented as raw single channel waveforms, speaker identification, and emotion labels encompassing various emotions such as anger, happiness, and sadness. It is worth noting that all the datasets exhibit similar average utterance durations, which range from 2.5 to 4.5 seconds.

\begin{table}[htbp]
  \centering
  \begin{tabular}{lcc}
    \toprule
    Model               & Layers & Training Data \\
    \midrule
    wav2vec2 Base        & 12   & 960 (1)           \\
    wav2vec2 Large       & 24   & 960 (1)           \\
    wav2vec2 XLSR 53     & 24   & 56,000 (53)       \\
    wav2vec2 XLSR 300M   & 24   & 436,000 (128)     \\
    wav2vec2 ASR Large   & 24   & 960 + 960   (1)   \\
    HuBERT Base          & 12   & 960  (1)          \\
    HuBERT Large         & 24   & 60,000 (1)        \\
    HuBERT ASR Large     & 24   & 60,000 + 960  (1) \\
    \bottomrule
  \end{tabular}
  \caption{Overview of Feature Extractor Models with Number of Layers and Training Data. Dataset size is in Hours and number of languages are mentioned inside parenthesis and '+' indicates additional finetuning training data hours.}
  \label{tab:models}
\end{table}

The selection of benchmark datasets for this study was primarily based on two key factors: dataset popularity and language diversity. The chosen benchmark datasets, including EmoDB \cite{EmoDB}, IEMOCAP \cite{IEMOCAP}, and RAVDESS \cite{RAVDESS}, are widely used in the field of speech emotion recognition. To address class imbalance, a subset of the IEMOCAP \cite{IEMOCAP} dataset, specifically the four emotion classes (IEM4), was used. For the remaining tasks, all samples and classes from the original datasets were retained. To represent Italic languages, CaFE \cite{CaFE} and EMOVO \cite{EMOVO} were chosen, while AESDD \cite{AESDD} and ShEMO \cite{ShEMO} represented the Hellenic and Indo-Iranian branches of the Indo-European family, respectively. The majority of the benchmark datasets primarily comprise scripted and acted speech, with IEM4, RAVDESS \cite{RAVDESS}, and ShEMO \cite{ShEMO} also incorporating spontaneous utterances.

To train, optimize, and evaluate language-specific speech emotion classifiers, each dataset, following Scheidwasser et al. \cite{scheidwasser2022serab}, was divided into training, validation, and testing sets. The standard split employed for most datasets involved allocating 60\% of the data for training, 20\% for validation, and 20\% for testing purposes. Speaker independence was carefully maintained in each partition, ensuring that the sets of speakers in each partition were mutually exclusive. This fixed data split design facilitated the assessment of the performance of experimental setups using different amounts of data, considering the variations in dataset sizes within the benchmark.

\section{Models}
\label{sec:models}

In this paper, we work with three pre-trained speech models - wav2vec2 \cite{baevski2020wav2vec}, XLSR \cite{conneau2020unsupervised} and HUBERT \cite{hsu2021hubert}. We use these models as feature extractors. A summary of these models can be found in Table \ref{tab:models}, presenting a concise overview of their key characteristics. For classification, we use to different heads on top of the features extracted from these speech representation models, as described in section \ref{subsec:heads}.

\subsection{Feature Extractors}\label{subsec:feature extractors}
We selected feature extraction models that vary in terms of pre-training data and the number of languages involved. By incorporating models with distinct pre-training data and linguistic coverage, we aimed to enhance the comprehensiveness and robustness of our research findings.

\subsubsection{wav2vec2}

A total of 3 versions of wav2vec2 were studied. Among them, two versions are pre-trained models, namely Wav2vec2 Base and Wav2vec2 Large. Additionally, there is one version of wav2vec2 finetuned for ASR called wav2vec2 ASR Large. These models differ in several aspects, including the number of layers in the transformer encoder and the training data hours, as well as the number of languages they were pre-trained on. These statistics are shown in Table \ref{tab:models}.

The pre-training of wav2vec2 base, wav2vec2 large, and wav2vec2 ASR large was performed using the LibriSpeech dataset, which consists of English speech data \cite{LibriSpeech}.

\subsubsection{XLS-R}

We studied two versions of the XLS-R model: Wav2vec2 XLSR 53 and Wav2vec2 XLSR 300M. These models share the same architecture, with 24 encoder layers and 300M parameters.

Wav2vec2 XLSR 53 was pre-trained on a diverse dataset consisting of 53 languages. On the other hand, Wav2vec2 XLSR 300M was pre-trained on an even more extensive dataset, comprising 128 languages. The inclusion of these two models allowed us to explore the impact of different language coverage on the performance and capabilities of the XLS-R model in our research.

\begin{figure}
  \centering
    \includegraphics[width=0.48\textwidth]{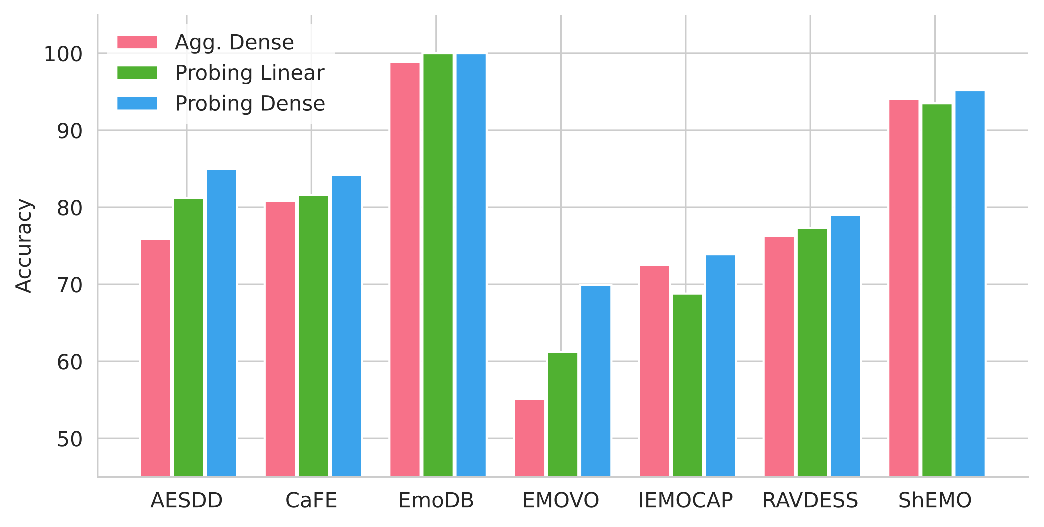}
  \caption{Maximum accuracies for all the classifiers models and datasets.}
  \label{fig:acc_comparison_plot}
\end{figure}

\subsubsection{HuBERT}

In our research experiments, we explored HuBERT by utilizing two pretrained models and one finetuned for ASR version. The pretrained versions of HuBERT include HuBERT Base, which has 12 encoder layers, and HuBERT Large, which has 24 encoder layers. HuBERT Base was pretrained on 960 hours of the LibriSpeech English dataset \cite{LibriSpeech}. On the other hand, HuBERT Large was pretrained using a significantly larger dataset, specifically 60,000 hours of the Libri-Light English dataset \cite{librilight}. We also employed a finetuned version of HuBERT for ASR, called HuBERT ASR Large. This model consists of 24 layers in its encoder and it was pretrained on 60,000 hours of the Libri-Light dataset and then further finetuned on 960 hours of the LibriSpeech dataset.

\subsection{Classification Heads}
\label{subsec:heads}
 Classification head is a model which takes in the learned features or representations of the input speech data from the preceding feature extractor and make predictions about the target class or emotion which a given input belongs. We used two classification heads which are described below.
 
\subsubsection{\textit{Linear}}
The \textit{Linear} classification head consists of two linear layers. First layer takes the averaged input features over the time sequence, and the rectified linear unit (ReLU) activation function is applied to the output with hidden dimension as 128. The resulting tensor is then passed through the second layer, which produces the final logits for each label. This classification head is based on the probing models used in \cite{shah2021all} \cite{pasad2021layer} \cite{pasad2023comparative}. The idea here is to use the simplest feed forward neural network to be able to study the speech representation power of pre-trained models.

\subsubsection{\textit{Dense}}
\textit{Dense} classification model uses piece-wise linear layers at every time step of a layer or a CNN layer with kernel size 1. Two such layers are used with the hidden dimensions of 256 hidden units. These features are then averaged across time and then passed through a final classification layer. This model is used with both single layer features and multilayer features. When using features from multiple layers, we aggregate the features following \cite{pepino2021emotion} before using the dense layer. 

\begin{figure}
  \centering
    \includegraphics[width=0.48\textwidth]{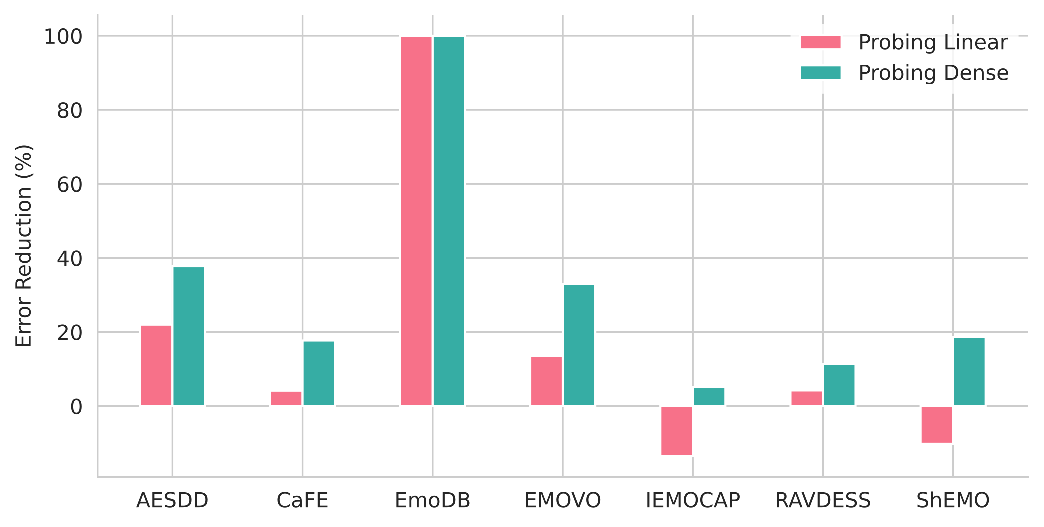}
  \caption{Error reduction percentage from Aggregate \textit{Dense} classifier to Probing with \textit{Linear} and \textit{Dense} classifier.}
  \label{fig:acc_pc_plot}
\end{figure}

\section{Experiments}
\label{sec:experiments}

We conducted two fundamental experiments by integrating all the feature extractors and classification heads. The first experiment involved a straightforward aggregation model, while the second experiment focused on edge probing to gain insights into the internal encoding schema of emotions in speech representation models. As part of our research, we treated the CNN layer that precedes the transformer layer as a separate layer, which we have denoted as the "zeroth" layer \cite{pepino2021emotion}. To enhance the statistical validity and ensure the consistency of our findings, we conducted each individual trial five times.

\begin{table*}[!t]
  \centering
  \begin{tabular}{l|ccccccc}
    \toprule
    \multicolumn{1}{l}{\textbf{Models}} & \multicolumn{7}{c}{\textbf{Datasets}} \\ 
    \cmidrule(lr){2-8}
    \textbf{}           & \textbf{AESDD}  & \textbf{CaFE}   & \textbf{EmoDB}   & \textbf{EMOVO}   & \textbf{IEMOCAP} & \textbf{RAVDESS} & \textbf{ShEMO}  \\
    \midrule   
    wav2vec 2 Base      & 69.6 [5]  & 70.9 [2]  & 91.7  [3]  & 44.4 [5]   & 66.8 [6]   & 65.3 [8]   & 92.2 [3]  \\
    wav2vec 2 Large     & 79.7 [15] & 79.9 [8]  & 98.8  [9]  & 50.5 [16]  & 67.7 [7]   & 73.7 [12]  & 91.9 [5]  \\
    wav2vec 2 ASR LARGE & 79.2 [8]  & 75.2 [12] & 97.6  [9]  & 56.1 [9]   & 69.0 [13]  & 70.3 [11]  & 92.2 [4]  \\
    wav2vec 2 XLSR 53   & \textbf{85.0 [14]} & \textbf{84.2 [7]}  & \textbf{100 [13]} & 61.7 [9]   & 71.7 [10]  & 77.7 [9]   & \textbf{95.2 [6]}  \\
    wav2vec 2 XLRS 300  & 83.1 [16] & 83.3 [13] & \textbf{100 [7]}  & \textbf{69.9 [14]}  & \textbf{73.9 [13]}  & \textbf{79.0 [18]}  & 94.6 [17] \\
    HuBERT Base         & 72.9 [7]  & 73.1 [7]  & 96.4  [4]  & 48.5 [6]   & 69.2 [4]   & 65.7 [7]   & 91.7 [1]  \\
    HuBERT Large        & 80.2 [14] & 83.8 [13] & \textbf{100 [15]} & 62.8 [12]  & 72.5 [22]  & 76.3 [11]  & 93.5 [5]  \\
    HuBERT ASR LARGE    & 82.6 [13] & 82.5 [10] & \textbf{100 [10]} & 67.3 [12]  & 71.4 [10]  & 75.3 [12]  & 93.8 [8]  \\
    \bottomrule
  \end{tabular}
  \caption{Maximum accuracies for Probing using the \textit{Dense} classifier with various feature extraction models. The corresponding layer of the encoder in the feature extractor is denoted within square brackets.}
  \label{tab:acc_prob_dense}
\end{table*}

\subsection{Edge Probing - Linear}
Edge probing involves conducting targeted analyses and experiments to investigate how the model's internal representations and attention mechanisms respond when an input is sent through the model. For a given model with layers L, we attached a separate classification head to each layer and trained them independently to predict a target label. The classifier solely relies on input from a single layer. We conducted this experiment using both a \textit{Linear} classification head and a \textit{Dense} classification head. Consequently, the classifier heavily relies on the encoder to provide meaningful information regarding the relationship between spans and their respective roles within a sentence. This approach empowered us to evaluate the encoder's proficiency in capturing and conveying emotional features from the high density speech input data.

The results of the probing experiments for five models are shown in Figure \ref{fig:linear_acc_plots}. We find that the initial and final layers perform worst for the task of speech emotion recognition. The initial layers are unable to create a rich enough representation of speech to classify emotions and classifying emotions requires both the understanding of the phonetic as well as prosodic content. The final layers of these models, as shown in \cite{shah2021all} \cite{pasad2021layer} \cite{pasad2023comparative} are focused on reconstructing the input and lose the rich contextual representations for emotion recognition to be able to provide enough phonetic content for speech reconstruction. This can also be seen in Figure \ref{fig:linear_acc_plots} where the performance of the final ASR layers are always worse than the non-ASR models, showing that models trained to convert speech to text lose out on a lot more prosodic information than their non-ASR counterparts. The center layers contain the richest contextual features that contain enough phonetic and prosodic content to do speech emotion recognition. This observation is true across six different languages.

\subsection{Aggregation}
The aggregation experiments follow the work in \cite{pepino2021emotion}, where feature representations from all the layers are used to train an SER model. To combine features from different layers, we use a weighted average for each layer as done in \cite{pepino2021emotion}, where these weights are learnable parameter. We then use the dense layer to work with these weighted multi-layer features. Upon performance evaluation of this model, we found that the aggregation based models were performing worse than the linear edge-probing models. This can also be seen in Figure \ref{fig:acc_comparison_plot}, where Agg. Dense performance if always worse than the Probing with\textit{Linear} classifier performance except for the IEMOCAP (English) dataset. This motivated us to push the performance using a single layer of the pre-trained models using the same classification head. 

\subsection{Edge Probing - Dense}
To extract even more out of the edge probing models, we use the dense classification head on the features extracted from a single layer. We find that the the dense probing models outperform both linear probing and dense aggregate models as shown in Figure \ref{fig:acc_comparison_plot}. This means that features extracted from a single layer are enough to achieve best performance on SER over using features extracted from all layers, even though the features are combined using learnable weights. 

The improvement in performance over aggregation experiments is highlighted further in Figure \ref{fig:acc_pc_plot}. The plots show the error reduction percentage over the aggregation models when using the linear and dense probing models. We observed that for 5 out of 7 datasets, the linear probing model performs better than the aggregation models, and the dense probing model is better than the aggregation model for all datasets. We also noticed that the dense probing model closes the error of the aggregation model by 5-100\%, with an average error reduction about 32\% across 7 datasets and 6 languages. 

Figure \ref{fig:acc_pc_plot} also shows that the maximum improvements with dense probing models happen for the smallest datasets. This means that for low resourced scenarios, if the training dataset is small, the middle layers become even more crucial for accurately classifying emotions and aggregation models as proposed in \cite{pepino2021emotion} require a larger amount of data to achieve optimal performance. The exact values of classification accuracies for the dense probing head with the corresponding best layer for each model are shown in Table \ref{tab:acc_prob_dense}. Unsurprisingly, we find the XLSR models trained on large multilingual data performs best for speech emotion recognition, including doing SER for English.  

\section{Conclusion}
\label{sec:conclusion}
We conducted a comprehensive evaluation of transformer-based speech representation models for speech emotion recognition (SER) across multiple languages. Our findings challenge prevailing notions and provide valuable insights for optimizing SER models in multilingual and low-resourced scenarios. Our probing experiments show that the middle layers of the speech representation models capture the most important features for SER. This is contrary to previous work \cite{lin2023utility}, which shows that the early layers of speech models capture prosodic information, highlighting the importance of both phonetic and prosodic information for recognizing emotion in speech. 

We also discovered that single-layer probing models consistently outperformed aggregation models in terms of accuracy and variability. The findings of this study challenge the previous work by Pepino et al. \cite{pepino2021emotion}, as they demonstrate that utilizing features from a single layer of a speech representation model outperforms the approach of aggregating features from all layers of the model. The optimal layers were usually the center layers of the model, thus further highlighting the importance of the center layers for the task of speech emotion recognition. The optimal single layer although seems to differ from task to task and dataset to dataset, and finding the optimal layer is a part of our future investigations. 

Additionally, dataset size played a crucial role, with aggregation models requiring more data to perform well. Furthermore, we observed that models trained on a larger number of languages exhibited better encoding of emotions, emphasizing the importance of linguistic diversity in pre-training. 

\bibliographystyle{IEEEbib}
\bibliography{refs}

\end{document}